\documentclass{article}

\PassOptionsToPackage{numbers, compress}{natbib}
\usepackage[preprint]{neurips_2020}

\usepackage[utf8]{inputenc} 
\usepackage[T1]{fontenc}    
\usepackage{hyperref}       
\usepackage{url}            
\usepackage{booktabs}       
\usepackage{amsfonts}       
\usepackage{nicefrac}       
\usepackage{microtype}      

\usepackage{times}

\usepackage{soul}
\usepackage[small]{caption}
\usepackage{graphicx}
\usepackage{algorithm}
\usepackage{algorithmicx}
\usepackage{algpseudocode}
\usepackage{multirow}
\usepackage{amsmath}
\usepackage{xcolor}
\title{ADD: Augmented Disentanglement Distillation Framework for Improving Stock Trend Forecasting}

\author{
	Hongshun Tang$^1$\thanks{This work was done when the author was an intern at Microsoft Research Asia.},
	Lijun Wu$^2$, 
	Weiqing Liu$^2$, 
	Jiang Bian$^2$\\
	$^1$Peking University\\
	$^2$Microsoft Research Asia\\
	1801210885@pku.edu.cn, \{Lijun.Wu, Weiqing.Liu, Jiang.Bian\}@microsoft.com
}

\begin{document}
	
\maketitle

\begin{abstract}
	Stock trend forecasting has become a popular research direction that attracts widespread attention in the financial field. Though deep learning methods have achieved promising results, there are still many limitations, for example, how to extract clean features from the raw stock data. In this paper, we introduce an \emph{Augmented Disentanglement Distillation (ADD)} approach to remove interferential features from the noised raw data. Specifically, we present 1) a disentanglement structure to separate excess and market information from the stock data to avoid the two factors disturbing each other's own prediction. Besides, by applying 2) a dynamic self-distillation method over the disentanglement framework, other implicit interference factors can also be removed. Further, thanks to the decoder module in our framework, 3) a novel strategy is proposed to augment the training samples based on the different excess and market features to improve performance. We conduct experiments on the Chinese stock market data. Results show that our method significantly improves the stock trend forecasting performances, as well as the actual investment income through backtesting, which strongly demonstrates the effectiveness of our approach.
\end{abstract}

\section{Introduction}

Recently, stock trend forecasting has become an important research topic in the financial field. In practice, the stock return can be divided into two parts, one is the market return, which represents the overall trend of the stock market, and the other one is the excess return, which displays the changes of each stock affected by its own factors outside the market changes \citep{mcwilliams1997event}. Both returns provide valuable information to benefit real investors to analyse the market trend and each specific stock so to help make investment decision. Therefore, predicting market return and excess return are both worthy in stock trend forecasting. 

To model the prediction of stock return, traditional methods include AutoRegressive \cite{li2016stock}, Moving Average \cite{metghalchi2012moving} and Kalman Filters \cite{bisoi2014hybrid}. Nowadays, the deep neural networks (DNNs) have shown great potential, which significantly outperforms the conventional approaches \cite{hu2018listening,li2019multi,chen2019investment}. Though promising results are obtained, the \emph{interferential features} existed in the noised raw stock data have seriously restricted the model performance \cite{feng2018enhancing,wei2016hybrid}. 
In this work, we focus on the market and excess return predictions, and we find the factors related to these two returns are highly entangled/correlated in the noised raw data, which makes the market factors be interferential features for the excess return prediction \cite{hadad2018two}. Similarly, the excess factors also hurt the performance of market return prediction. Besides, other noisy interferential factors also disrupt the two return predictions. Therefore, it is critical to remove the interferential features so that both return predictions can be benefited. 

In order to decouple the entanglement between the market and excess factors, we present an Augmented Disentanglement Distillation (ADD) approach, where 1) a \emph{disentanglement framework} to separate the market and excess information, and 2) a \emph{dynamic self-distillation} method to enhance the related return features. Also, 3) a novel \emph{data augmentation} strategy is introduced, which can greatly improve the performance for rare data samples. Specifically, our disentanglement model consists of two separated encoders, two return predictors and a reconstructive decoder. The market and excess features are extracted by their respective encoders. By utilizing the corresponding predictors (which predict the market/excess returns using market/excess information), and two extra predictors with adversarial training (which prevent the market/excess features predicting the excess/market returns), the market and excess factors can be disentangled. 

The self-distillation training method is leveraged to further enhance the clean knowledge for market/excess features. Different from conventional knowledge distillation, we dynamically distill the knowledge from the teacher model with weights calculated according to the teacher performance, so to better control the contribution from the teacher knowledge and raw data during student model training.

Thanks to the reconstructive decoder and separated encoders, we augment the stock training data in a novel way. Concretely, we feed the disentangled features outputted by market/excess encoders from different original samples into decoder, so to get the augmented new samples. In practice, we focus more on augmenting those rare/hard samples.

We conduct experiments on the Chinese stock market of A-shares from $2007$ to $2019$. In addition, we also add a trading strategy backtesting experiment to evaluate our method in real investment market. The strong results of the excess and market return predictions demonstrate the effectiveness of our method, and the backtesting results further prove that our approach is beneficial for real investment. 

In summary, our main contributions are as follows:
\begin{itemize}
	\item
	We propose a disentanglement structure to separate the excess and market information and remove the entangled relationship between them.
	\item
	We develop a dynamic self-distillation method through iterative training to further enhance the return features.
	\item
	We introduce a novel augmentation method based on the disentanglement features to improve model performance, especially for the rare samples.
	\item
	Experiments show that our Augmented Disentanglement Distillation (ADD) method can improve the stock trend forecasting and increase the investment income.
\end{itemize}

\section{Method}
\subsection{Background}
\label{sec:background}

Before discussing our method, we first introduce the background and necessary notations.

Stock return is often defined as the future change rate of stock price, which is formulated as:
\begin{align}
r_i^j = \frac{price_i^{j+1}-price_i^j}{price_i^j},
\end{align}
where $r_i^j$ and $price_i^j$ are the stock return and price of stock $s_i$ on day $d_j$. The price can be opening price, closing price or Volume Weighted Average Price, we choose closing price in the work.
The daily market return is the average of the stock returns of all stocks on a certain day, which represents the overall trend of stock market. The excess return is the difference between the stock return and market return, which represents the changes of each stock outside the market changes.

We formalize the task as follows: $X$ denotes the stock features (see Section \ref{sec:data}), $Y_E$ and $Y_M$ are the labels of excess and market returns. Given a dataset $D=\{(x, y_e, y_m)\}, x \in X, y_e \in Y_E, y_m \in Y_M$, the goal is to predict the excess returns $Y_E$ and the market returns $Y_M$ based on $X$.

The market return for all stocks is same in one day, which makes the number of the market return values be far smaller than stock samples. Therefore, we perform a classification task for the market return. We divide all market returns into three categories by the thresholds calculated on the market distribution from training set. The thresholds fulfill that each category holding the same number of training samples, and they are also reused in validation and test sets. The three categories are the market return `going up', `moving steady' and `going down'. As for the excess prediction, we carry on the regression task. 

\subsection{Disentanglement Framework}

\begin{figure}[!tbp]
	\centering
	\includegraphics[width=0.6\linewidth]{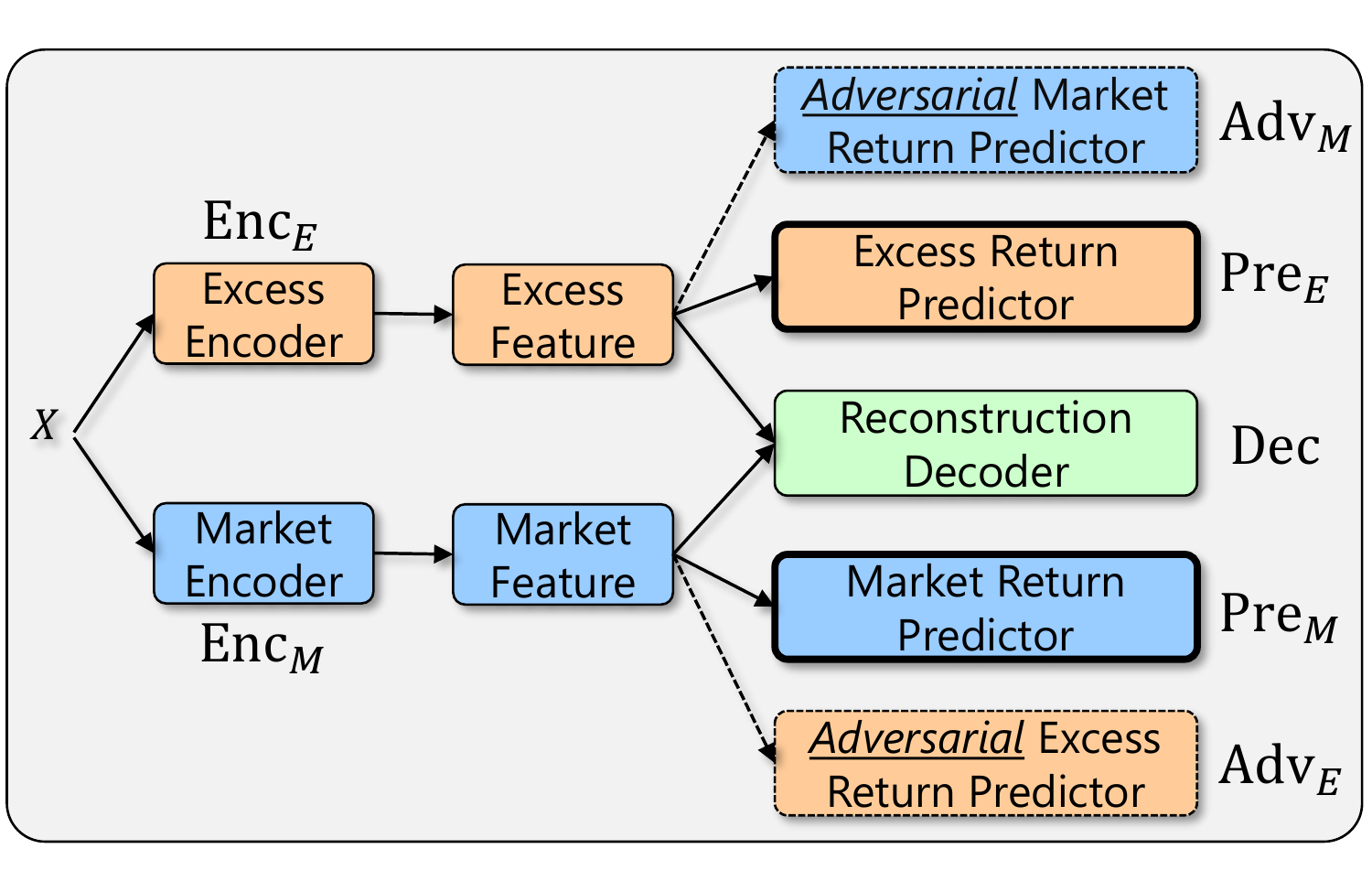}
	\caption{The overall disentanglement framework, which contains the excess and market return predictors, the two corresponding adversarial predictors and the reconstruction decoder.}
	\label{fig:distenglement_framework}
\end{figure}

As discussed before, the market factors are interferential features for excess return prediction, and vice versa. To separate the market and excess factors for accurate predictions, we propose a disentanglement framework (shown in Figure \ref{fig:distenglement_framework}). 
The framework consists of two encoders, one decoder and four predictors. Specifically, we use excess encoder \texttt{Enc}$_E$ and market encoder \texttt{Enc}$_M$ to encode stock features $X$ into excess feature $f_E = \texttt{Enc}_E(X)$ and market feature $f_M = \texttt{Enc}_M(X)$. The decoder \texttt{Dec} takes the disentangled $f_E$ and $f_M$ as inputs and outputs $\widehat{X}$, which is to recover the original inputs $X$. The decoder is designed to reduce the loss of information during encoding process, which plays an important role for later augmentation (Section \ref{sec:augmentation}). 

To keep the features $f_E$ and $f_M$ only related to its own return, we build two predictors for each disentangled feature, one for preserved information (e.g., excess return predictor \texttt{Pre}$_E$) and the other one for unrelated information (e.g., adversarial predictor \texttt{Adv}$_M$). The objective is to minimize the sum of three weighted terms: 1) the loss of preserved information, 2) the negative adversarial loss of unrelated information, and 3) the reconstruction loss of decoder.

Formally, let $\theta_{Enc}$ be the parameters of \texttt{Enc}$_E$ and \texttt{Enc}$_M$,  $\theta_{Dec}$ be the parameters of \texttt{Dec}, $\theta_{Pre}$ be the parameters of \texttt{Pre}$_E$ and \texttt{Pre}$_M$, $\theta_{Adv}$ be the parameters of \texttt{Adv}$_E$ and \texttt{Adv}$_M$. We define $\mathcal{L}_{Pre}$ as the prediction losses for \texttt{Pre}$_E$ and \texttt{Pre}$_M$, $\mathcal{L}_{Adv}$ for \texttt{Adv}$_E$ and \texttt{Adv}$_M$, $\mathcal{L}_{Rec}$ for \texttt{Dec}. 
As for optimization, the objective is to simultaneously minimize $\mathcal{L}_{Pre}$ and $\mathcal{L}_{Rec}$ while maximize $\mathcal{L}_{Adv}$. $\lambda$ and $\mu$ are used to weight the losses. The two training objectives are executed alternately, just like the general process of adversarial training. Assume that \texttt{MSE} represents mean square error, and \texttt{CE} represents cross entropy, the whole training loss is:
\begin{align*}
\mathcal{L}_{Pre} &= \texttt{MSE}(\texttt{Pre}_E(f_E), Y_E) + \texttt{CE}(\texttt{Pre}_M(f_M), Y_M), \\
\mathcal{L}_{Adv} &= \texttt{MSE}(\texttt{Adv}_E(f_M), Y_E) + \texttt{CE}(\texttt{Adv}_M(f_E), Y_M), \\
\mathcal{L}_{Rec} &= \texttt{MSE}(X, \widehat{X}).
\end{align*}
\begin{align}
\label{eqn:disentangle_loss}
\mathop{min}\limits_{\theta_{Enc},\theta_{Pre},\theta_{Dec}} \mathcal{L}_1 &= \mathcal{L}_{Pre} - \lambda * \mathcal{L}_{Adv} + \mu * \mathcal{L}_{Rec},  \nonumber \\
\mathop{min}\limits_{\theta_{Adv}} \mathcal{L}_2 &= \mathcal{L}_{Adv}.
\end{align}

We choose GRU \cite{chung2014empirical} network for our encoder and decoder. The last output of GRU encoder will feed into a MLP layer for the disentangled features. Similarly, the decoder takes the excess and market features as input and outputs the recovered $\widehat{X}$ through another MLP layer. 

\subsection{Dynamic Self-Distillation}
\label{sec:distillation}

Though the market and excess factors can be separated by return predictors and adversarial predictors, there may still exist unrelated factors which influence the prediction performance. To further remove the interferential features and enhance the representation, we incorporate a self-distillation \cite{zhang2019your,kim2020self} method to distill the disentangled feature information from the teacher model to student model, where the teacher and student models are the same disentanglement networks in our work. 

Different from the conventional distillation, we introduce a dynamic self-distillation method, where the knowledge weight for a sample is determined by its own performance and the corresponding day performance from the teacher model. The inspiration is to dynamically control the knowledge importance for different samples. For example, if the teacher performance is not good on a day, we should increase the weight for all samples on that day. Besides, if one sample's performance is not good, we increase the weight for that sample. Otherwise, we reduce the weight of these samples.

To measure excess return performance, Information Coefficient (\texttt{IC}) is used for day performance and Mean Squared Error (\texttt{MSE}) loss for each sample. Let $ic_{max}$ and $ic_{min}$ be the maximum and minimum \texttt{IC} values among all days, $mse_{max}$ and $mse_{min}$ be the maximum and minimum \texttt{MSE} values among all samples. $ic^j$ is the \texttt{IC} for day $d_j$ and $mse_i^j$ is the \texttt{MSE} for stock sample $s_i$ on day $d_j$. We define two biases $\beta_{day}$ and $\beta_{sample}$, which are used to ensure the importance of model knowledge. $\alpha$ is used for leveraging the influence of day and sample information. The final knowledge weight $w_i^j$ for stock sample $s_i$ on day $d_j$ is formulated by the daily weight $wd^j$ and the sample weight $ws_i^j$ as follows:
\begin{align}
\label{eqn:dynamic}
wd^j &= \beta_{day} + (1 - \beta_{day}) * \frac{ic_{max} - ic^j} {ic_{max} - ic_{min}},  \nonumber\\
ws_i^j &= \beta_{sample} + (1 - \beta_{sample}) * \frac{mse_{max} - mse_i^j} {mse_{max} - mse_{min}},  \nonumber \\
w_i^j &= \alpha * wd^j + (1 - \alpha)  * ws_i^j.
\end{align}

In this way, we can build the self-distillation loss for encoders between teacher and student models. Let $ht_{i}^{j}$ and $hs_{i}^{j}$ represent the last step output of teacher model and student model respectively. The distillation loss is:
\begin{align}
\mathcal{L}_{Dis} = \sum_{i} \sum_{j} w_i^j * \texttt{MSE}(ht_{i}^{j}, hs_{i}^{j}).
\end{align}
The distillation for market encoder is same as excess encoder after replacing with the classification measure in Eqn. (\ref{eqn:dynamic}).

Define the parameter $\xi$ to control the weight of the distillation loss. Combined with previous disentanglement training, the overall training objectives are:
\begin{align}
\label{eqn:final_loss}
\mathop{min}\limits_{\theta_{Enc},\theta_{Pre},\theta_{Dec}} \mathcal{L}_1 &= \mathcal{L}_{Pre} - \lambda * \mathcal{L}_{Adv} + \mu * \mathcal{L}_{Rec} + \xi * \mathcal{L}_{Dis},  \nonumber \\
\mathop{min}\limits_{\theta_{Adv}} \mathcal{L}_2 &= \mathcal{L}_{Adv}.
\end{align}

\subsection{Data Augmentation}
\label{sec:augmentation}

\begin{figure}[!tbp]
	\centering
	\includegraphics[width=0.6\linewidth]{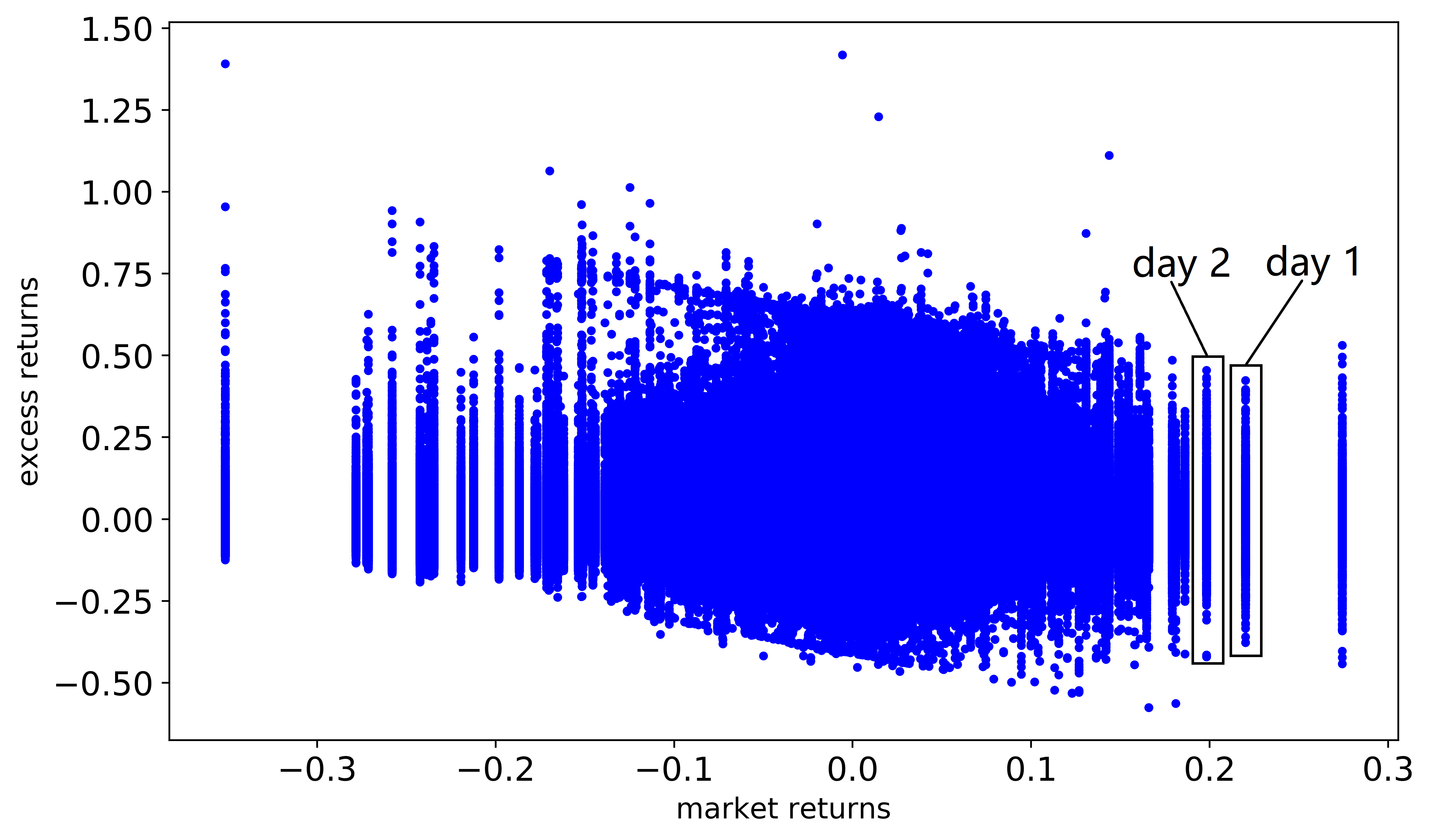}
	\caption{The distribution of stock returns. As an example, to augment the training data samples, we choose the samples in day 1 and day 2 which have close market returns.}
	\label{fig:distribution_return}
\end{figure}

To further improve the model generalization, we propose a novel data augmentation method to extend the stock training data, where the new data samples are generated by the decoder with disentangled features from different samples. 

For a sample $p = \{x^p, y_e^p, y_m^p$\} and another one $q =\{x^q, y_e^q, y_m^q$\}, we obtain a new sample based on the excess feature $f_{E}^p$ from $p$ and the market feature $f_{M}^q$ from $q$. $f_{E}^p$ and $f_{M}^q$ are then concatenated and fed into the decoder to generate $\widehat{x}$. Therefore, the new sample is augmented as \{$\widehat{x}, y_e^p, y_m^q$\}. Similarly, we can use the excess feature $f_{E}^q$ from $q$ and the market feature $f_{M}^p$ from $p$ to generate another new sample. 

Now the question is how to choose samples $p$ and $q$. 
In principle, the disentangled features should be taken from the two days that the market returns are close. As shown in Figure \ref{fig:distribution_return}, the distributions of excess and market returns are close to each other. Therefore, if we choose from the days that market returns are close, the augmented samples should also be close to the original data. 
In addition, we pay more attention to the days that perform not good enough (the days that \texttt{IC} measurement is low), so to improve the performance of these rare and hard samples.

\section{Training Algorithm}
\begin{algorithm}[!htb]  
	\caption{ADD Algorithm}
	\scalebox{0.9}{
		\begin{minipage}{0.9\linewidth}
			\begin{algorithmic}[1]
				\Require Disentanglement model $M$,
				training dataset $D$
				\State Train model $M$ on data $D$ by losses Eqn. (\ref{eqn:disentangle_loss})
				\While{$M$ not converge}
				\State Set teacher model $M_t = M$
				\State Generate new samples dataset $\widehat{D}$ (\emph{Section \ref{sec:augmentation}})
				\State Augment the training data $D = D \cup \widehat{D}$
				\State Train new model $M$ on data $D$ by losses Eqn. (\ref{eqn:final_loss}) with knowledge distilled from $M_t$ (\emph{Section \ref{sec:distillation}})
				\EndWhile
				\State Return model $M$
			\end{algorithmic} 
		\end{minipage}
	}
	\label{alg:add}
\end{algorithm}

The proposed Augmented Disentanglement Distillation (ADD) method can be seen in Algorithm \ref{alg:add}. 

At first, we train the disentanglement model with excess and market return predictors, and the adversarial predictors. The trained disentanglement model will serve as the teacher model later. After that, we dynamically augment new samples (Section \ref{sec:augmentation}) and take augmented ones together with raw samples as the new training dataset, then we distill the knowledge from the teacher model (Section \ref{sec:distillation}) and train a new student model for return predictions. This student model will server as teacher model in the next round, and this training procedure can be iteratively performed until convergence. 

\section{Experiment}

We conduct experiments on the Chinese stock market for both stock return predictions and the trading strategy backtesting. The return prediction task can reflect the predicting ability of our method while the backtesting result can demonstrate the actual effect in the real investment application.

\subsection{Data}
\label{sec:data}
The data source is China A-shares stock market, which is one of the fastest growing stock markets in the world. We obtained $13$ years of stock data from $2007$ to $2019$, with a total of $7,836,559$ samples. The data is split as follows: the training set consists of $3,078,000$ data samples from $2007$ to $2013$, the validation set contains $1,114,234$ samples from $2014$ to $2015$ and $3,062,042$ data samples for the test set from $2016$ to $2019$. We take three years of test data, which is enough to verify the stability of the method in time.

Each sample contains $360$ ($6\times60$) dimensional volume and price features, and the labels of the excess and the market returns. The format of the data sample is introduced as in Section \ref{sec:background}.  The features are composed of $6$ factors that reflect the important direction of the stocks, namely opening price, closing price, highest price, lowest price, volume and Volume Weighted Average Price (VWAP). Each factor takes the corresponding values of the past $60$ days. Therefore, the features contain valuable time series information.

\subsection{Setting}

For the disentanglement model, we adopt the one-layer GRU \cite{chung2014empirical} for the encoder and decoder networks with hidden size $64$. For the four predictors, we leverage the $2$-layer stacked \texttt{MLP} network with batch normalization \cite{ioffe2015batch}, the activation function is \texttt{tanh}. As introduced, the market return prediction is a classification task while the excess return prediction is a regression task. Hence, one three class \texttt{softmax} layer is used for market return predictors, while one unit regression for excess return predictors. The regression loss is \texttt{MSE} for excess returns and the classification loss is cross entropy for market returns.
We use Adam \cite{kingma2014adam} optimizer with the learning rate $0.001$ for training. Dropout \cite{srivastava2014dropout} is used for the decoder output with value $0.5$, the batch size is $5,000$.

\begin{table}[!htb]
	\centering
	\scalebox{0.85}{
		\begin{tabular}{c c c}
			\toprule
			\textbf{Hyper} & \textbf{Represent} & \textbf{Value} \\
			\midrule
			$\lambda$ & weight of adversarial loss & $0.4$  \\
			$\mu$ & weight of reconstruction loss & $0.05$ \\
			$\xi$ & weight of distillation loss & $0.8$  \\
			$\beta_{day}$ & bias of daily weight &$0.5$ \\
			$\beta_{sample}$  & bias of sample weight & $0.2$ \\
			$\alpha$  & influence of daily and sample weight & $0.4$ \\
			\bottomrule
		\end{tabular}
	}
	\caption{Hyper-parameter settings used in our ADD method.}
	\label{tab:abs_hyper}
\end{table}

For the hyper-parameters, we report their settings and the descriptions in Table \ref{tab:abs_hyper}. Note that we also show the impact of these hyper-parameters in Section \ref{sec:hyper_impacts}.

\subsection{Evaluation Metrics}

In quantitative investment field, there are some unique indicators for evaluating the performance of stock prediction. Concretely, for excess returns, people often take Information Coefficient (\texttt{IC}) and \texttt{Rank IC} \cite{mcwilliams1997event} as evaluation metrics. \texttt{IC} stands for the Pearson Correlation Coefficient of the predicted sequence and the labeled sequence, while \texttt{Rank IC} is the Spearman's Rank Correlation Coefficient of two sequences. Both \texttt{IC} and \texttt{Rank IC} can show how closely the stock predictions match the stock results.

In addition, the classification performance of market return within each category is also considered. We take macro \texttt{F1} score, which is the harmonic mean of the precision and recall, as another metric to evaluate the performance in each class.

\subsection{Compared Method}
To demonstrate the effectiveness of our method, we make comparison with previous baseline methods, which include:

\textbf{Long Short-Term Memory (LSTM):} As an effective model used for time series data, LSTM \cite{hochreiter1997long} has been widely utilized in stock prediction \cite{chen2015lstm,nelson2017stock} that achieves great performance.

\textbf{Gated Recurrent Unit (GRU):} GRU \cite{chung2014empirical} is another version of recurrent neural network, but has fewer parameters than LSTM. It has also been widely used for stock prediction \cite{minh2018deep,xu2018stock}. This GRU baseline is same as the encoder network in our framework, but our method incorporates other components and training strategies instead of directly predicting the returns. 

\textbf{The two-step disentanglement method:} \cite{hadad2018two} present a two-step method for return predictions. It was the first time that the disentanglement model was introduced to stock prediction, which also separates the market and the excess information. However, their model structure and training method are quite different from ours.

\begin{table*}[!tbp]
	\centering
	\begin{tabular} {l c c c c} 
		\toprule
		\multirow{2}{*}{\textbf{Method}}&
		\multicolumn{2}{c}{\textbf{\texttt{Rank IC}}} & 
		\multicolumn{2}{c}{\textbf{\texttt{IC}}}\\
		\cmidrule{2-5}
		& \textbf{Pre\_E$\uparrow$} & \textbf{Adv\_E$\downarrow$} & \textbf{Pre\_E$\uparrow$} & \textbf{Adv\_E$\downarrow$} \\
		\midrule
		LSTM & $0.1348$ & $-$ & $0.1350$  & $-$ \\
		GRU & $0.1353$ & $-$ & $0.1343$ & $-$ \\
		\midrule
		Two-step & $0.1375$ & $0.0627$ & $0.1362$ & $0.0619$ \\
		\midrule
		\textbf{ADD (Ours)} & $\bf{0.1480}$ & $\bf{0.0306}$ & $\bf{0.1435}$ & $\bf{0.0275}$ \\
		\midrule
		\hspace{0.15cm}$-$Distill and Augment & $0.1396$ & $0.0354$ & $0.1380$ & $0.0353$ \\
		\hspace{0.15cm}$-$Distill & $0.1409$ & $0.0352$ & $0.1391$ & $0.0348$ \\
		\hspace{0.4cm}$+$Static Distill & $0.1437$ & $0.0339$ & $0.1408$ & $0.0323$ \\
		\hspace{0.15cm}$-$Augment & $0.1460$ & $0.0307$ & $0.1419$ & $0.0279$ \\
		\hspace{0.4cm}$+$Noise Augment & $0.1468$ & $0.0310$ & $0.1424$ & $0.0277$ \\
		\bottomrule
	\end{tabular}
	\caption{Performances of excess return predictor $\texttt{Pre}_E$ by excess encoder $\texttt{Enc}_E$ and adversarial excess return predictor $\texttt{Adv}_E$ by market encoder $\texttt{Enc}_M$.}
	\label{tab:excess_prediction}
\end{table*}

\subsection{Prediction Performance}
\texttt{}
We report the return prediction performances on test set for our approach and baseline methods in this subsection. 

First, the performances of excess return predictor \texttt{Pre}$_E$ and adversarial predictor \texttt{Adv}$_E$ are shown in Table \ref{tab:excess_prediction}. For LSTM and GRU baselines, they are only used to predict the excess returns. As shown, our ADD method effectively facilitates the ability of predicting excess return with remarkable promotion of \texttt{Rank IC} and \texttt{IC} (e.g., $0.1480$ v.s. $0.1375$ \texttt{Rank IC}). For adversarial prediction, our method successfully removes most of the excess information from the market encoder \texttt{Enc}$_M$, which results a significant decline in the excess return prediction. For example, the \texttt{Rank IC} for ours is $0.0306$ while $0.0627$ of two-step work \cite{hadad2018two}.

Second, the market return prediction of \texttt{Pre}$_M$ and adversarial prediction \texttt{Adv}$_M$ is shown in Table \ref{tab:market_prediction}. 
Again, we show strong improvements of the accuracy and \texttt{F1} score over the baseline methods. As expected, the adversarial predictor achieves lower accuracy and \texttt{F1} score than the two-step method \cite{hadad2018two}, which means the market feature has been removed more from the excess encoder \texttt{Enc}$_E$. 

In a short summary, our ADD approach achieves satisfactory results on the excess and market return predictions, and surpasses the previous works in a large margin. 

\begin{table*}[!tbp]
	\centering
	\begin{tabular} {l c c c c } 
		\toprule
		\multirow{2}{*}{\textbf{Method}}&
		\multicolumn{2}{c}{\textbf{Accuracy}} & 
		\multicolumn{2}{c}{\textbf{\texttt{F1} score}}\\
		\cmidrule{2-5}
		& \textbf{Pre$_M\uparrow$} & \textbf{Adv$_M\downarrow$} & \textbf{Pre$_M\uparrow$} & \textbf{Adv$_M\downarrow$}\\
		\midrule
		LSTM & $42.05\%$ & $-$ & $0.4172$ & $-$ \\
		GRU & $42.56\%$ & $-$ & $0.4231$ & $-$ \\
		\midrule
		Two-step & $42.67\%$ & $35.49\%$  & $0.4235$ & $0.3543$ \\
		\midrule
		\textbf{ADD (Ours)} & $\bf{46.35\%}$ & $\bf{34.36\%}$ & $\bf{0.4628}$ & $\bf{0.3435}$ \\
		\midrule
		\hspace{0.15cm}$-$Distill and Augment & $44.51\%$ & $35.18\%$ & $0.4491$ & $0.3520$ \\
		\hspace{0.15cm}$-$Distill & $44.62\%$ & $35.07\%$ & $0.4495$ & $0.3518$  \\
		\hspace{0.4cm}$+$Static Distill & $45.13\%$ & $34.87\%$ & $0.4510$ & $0.3501$ \\
		\hspace{0.15cm}$-$Augment & $46.05\%$ & $34.46\%$ & $0.4602$ & $0.3439$  \\
		\hspace{0.4cm}$+$Noise Augment & $46.14\%$ & $34.50\%$ & $0.4604$ & $0.3441$  \\
		\bottomrule
	\end{tabular}
	\caption{Performances of market return predictor \texttt{Pre}$_M$ by market encoder \texttt{Enc}$_M$ and adversarial market return predictor \texttt{Adv}$_M$ by excess encoder \texttt{Enc}$_E$.}
	\label{tab:market_prediction}
\end{table*}

\subsection{Ablation Studies}

As introduced, our approach consists of a disentanglement framework, a dynamic self-distillation and a novel data augmentation strategy. In order to reflect the importance of each component, we conduct ablation studies in this section. The detailed settings are as follows:

\textbf{1. ADD without distillation and augmentation:} We remove the distillation and augmentation components in ADD, only remain the disentanglement model. Compared with \cite{hadad2018two}, this setting can illustrate the advantage of our structure for disentanglement modeling. 

\textbf{2. ADD without distillation:} We perform ADD method without distillation, so as to identify the value of training with self-distillation. 

\textbf{3. ADD with static distillation:} We replace the dynamic self-distillation part in our ADD method with a static distillation, which directly transfers the knowledge from teacher to student model for each sample, without considering any day performance or controlled weights.

\textbf{4. ADD without augmentation:} We remove the augmentation part in ADD method, which can verify the contribution of our novel augmentation strategy.

\textbf{5. ADD with noise augmentation:} We replace the augmentation part with a simple Gaussian noise $N(0, 0.25)$ to the original data samples as new ones, to prove our disentanglement based augmentation approach is effective.

The results for above ablation studies are presented in the last several lines of Table \ref{tab:excess_prediction} and Table \ref{tab:market_prediction}. From these numbers, we can observe several findings: 
\begin{itemize}
	\item Our \emph{disentanglement} network design is more suitable for stock trend prediction (two-step method v.s. setting 1). 
	\item Both distillation and augmentation benefit the return predictions (ADD v.s. setting 1).  
	\item Our novel augmentation method based on disentangled features is more effective than the simple noise augmentation (setting 4 and 5).  
	\item Our distillation part contributes more than the augmentation in the ADD framework (setting 2 and 4).
	\item The \emph{dynamic} self-distillation is sorely important than the static distillation (setting 2 and 3). 
\end{itemize}

\subsection{Impact of Augmentation Data}

To give a more detailed analysis about data augmentation strategy, we investigate the impact from the different number of the augmented samples to the performance. The results on valid set are clarified in Table \ref{tab:Aug_parameters}. 

As expected, as we increase the number of augmented samples, the model performance is enhanced until reaching a threshold, then the performance starts to drop. Therefore, we use $1.0$ million augmented samples in our experiment. 

\begin{table}[!htbp]
	\centering
	\begin{tabular} {c c c c c} 
		\toprule
		\textbf{\#Augment} & \textbf{\texttt{Rank IC}} & \textbf{\texttt{IC}} & \textbf{Accuracy} & \textbf{\texttt{F1} score} \\
		\midrule
		$0$ & $0.1701$ & $0.1670$ & $47.40\%$ & $0.4735$ \\
		$250,000$ & $0.1703$ & $0.1674$ & $47.43\%$ & $0.4736$ \\
		$500,000$ & $0.1711$ & $0.1680$ & $47.55\%$ & $0.4749$\\
		$\bf{1,000,000}$ & $\bf{0.1720}$ & $\bf{0.1691}$ & $\bf{47.69\%}$ & $\bf{0.4758}$ \\
		$1,500,000$ & $0.1717$ & $0.1690$ & $47.60\%$ & $0.4753$ \\
		\bottomrule
	\end{tabular}
	\caption{Impact of the different number of augmented data samples on valid set.}
	\label{tab:Aug_parameters}
\end{table}

\subsection{Impact of the Hyper-parameters}
\label{sec:hyper_impacts}
For the sake of studying the hyper-parameters, we perform experiments with multiple values of the hyper-parameters. We investigate the weight $\lambda$ of adversarial loss, $\xi$ of distillation loss and $\alpha$ of knowledge weight, others differ little in our preliminary attempts. Specifically, we vary $\xi$ in $[0.6, 0.8, 1.0]$, $\alpha$ in $[0.0, 0.2, 0.4, 0.6, 0.8, 1.0]$ and $\lambda$ in $[0.2, 0.4, 0.6]$ and show the varied \texttt{Rank IC} performances.

\begin{figure}[!tbp]
	\centering
	\includegraphics[width=0.6\linewidth]{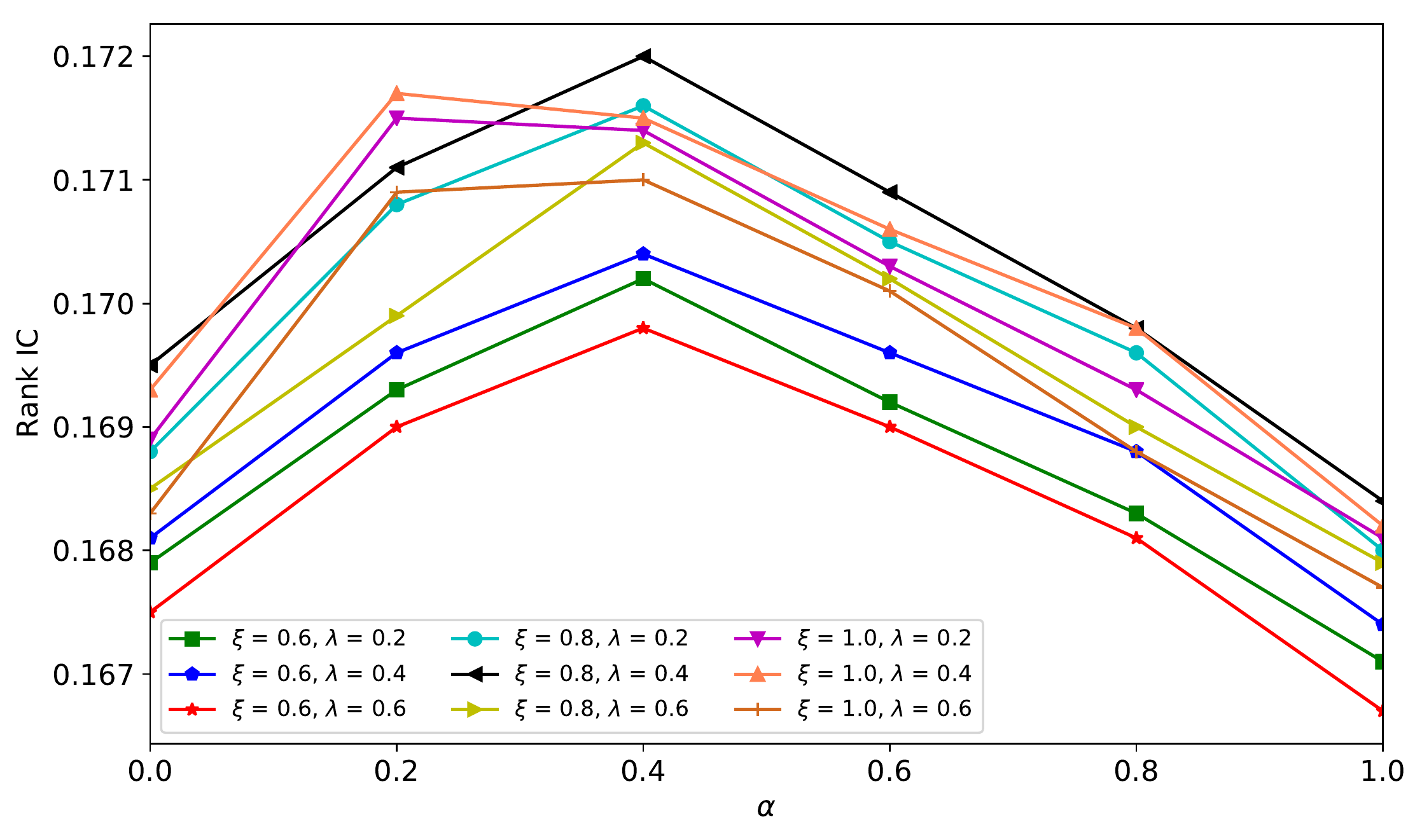}
	\caption{The performance for different hyper-parameter settings on validation set.}
	\label{fig_hyper}
\end{figure}

The influenced results on valid set are shown in Figure \ref{fig_hyper}. It is obvious that these hyper-parameters have a huge impact to the prediction performances. From these searched results, $\xi=0.8$, $\alpha=0.4$ and $\lambda=0.4$ make a best balance and contribution towards the final performance. Note that $\alpha=0.0$ and $\alpha=1.0$ mean that we use a single property instead of the combination of daily performance and sample performance, this result also demonstrates the necessary of the balanced combination process.

\subsection{Improvements for Rare Sample Performance}
Among all the samples, we pay more attention to those hard and rare examples. To evaluate the performance on rare samples, we first separate the test set into three subsets according to the performance of baseline model. Specifically, take the \texttt{Rank IC} for excess return and the probability value of the correct class for market return from baseline model as the division indicators, the test set is equally divided into three subsets: low set, middle set, and high set, with the same number of samples in each subset. Samples in the low set refer to the data that performs worst among the whole data set, while samples in the high set achieve the best performance. After splitting the test set, we then compare the performance of the baseline model and our method on each subset, and calculate the ratio of samples that ADD achieves better in each subset. The result is given in Table \ref{tab:Rare}.

\begin{table}[!htbp]
	\centering
	\begin{tabular}{l c c }
		\toprule
		\multirow{2}{*}{\textbf{Test Subset}}&
		\multicolumn{2}{c}{\textbf{Ratio of Samples}} \\
		\cmidrule{2-3}
		& \textbf{Excess} & \textbf{Market} \\
		\midrule
		Low Set & $78.77\%$ & $62.46\%$ \\
		Middle Set & $70.46\% $ & $56.62\%$ \\
		High Set & $61.23\%$ & $52.91\%$ \\
		\bottomrule
	\end{tabular}
	\caption{The ratio of data samples that our ADD method achieves better on the separated three subsets.}
	\label{tab:Rare}
\end{table}

We can see our ADD achieves better than the baseline model among all subsets with a large number of samples, especially on the low set (e.g., $78.77\%$ samples for excess return). Compared with middle set and high set, there are more samples achieving better performance in the low set than other two subsets (e.g., $78.77\%$ on low set v.s. $61.23\%$ on high set for excess return). These results demonstrate our ADD indeed cares more about the rare samples and improves their performances significantly. 


\subsection{Backtesting}

\begin{figure}[!tbp]
	\centering
	\includegraphics[width=0.6\linewidth]{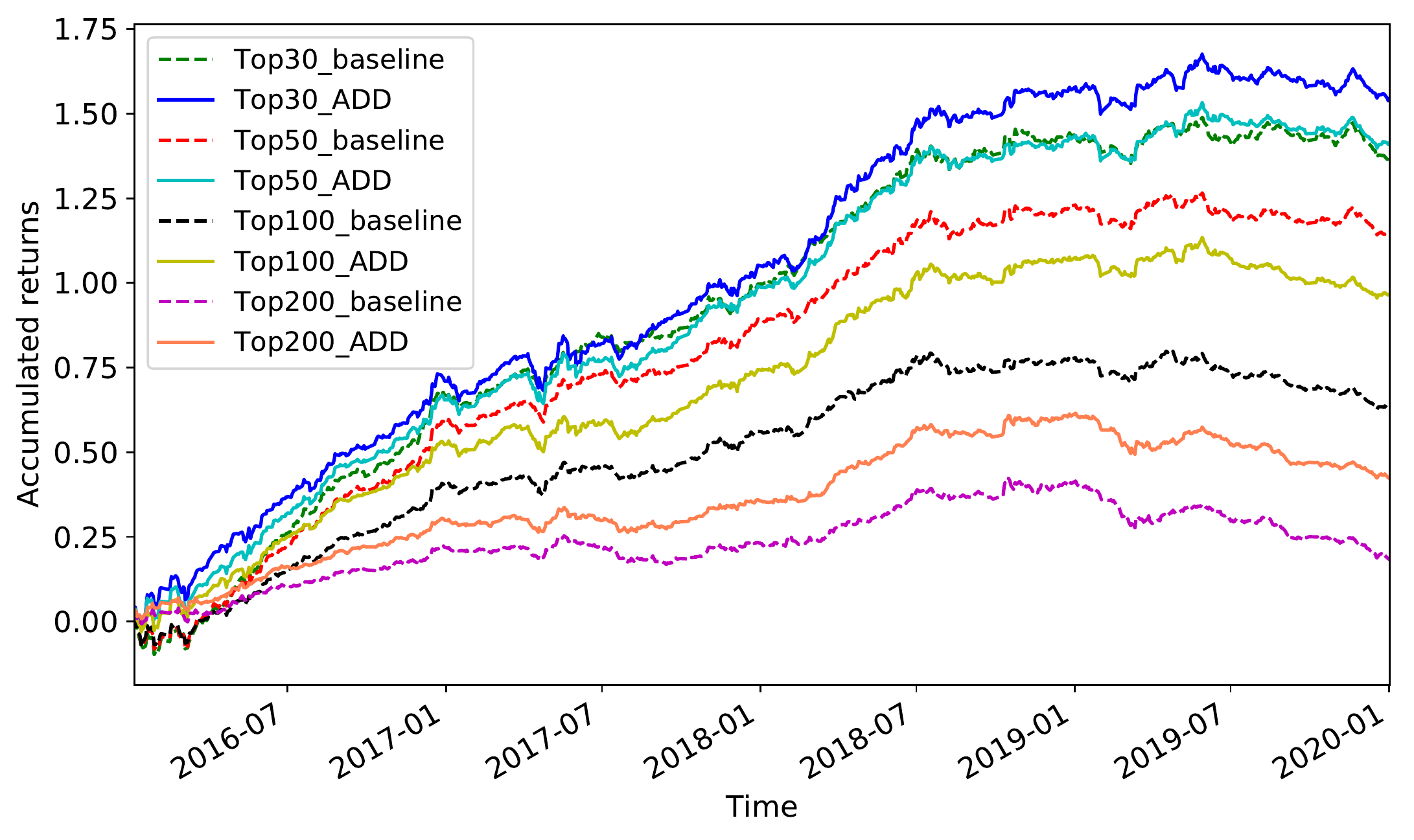}
	\caption{The accumulated returns over time in backtesting experiments.}
	\label{fig6}
\end{figure}

For verifying the role of our method in actual investment, we need to assess the viability of a trading strategy based on our prediction by discovering how it play out using real-world data, which is the idea of backtesting.

A widely-acknowledged method for backtesting is to pick the top-$k$ stocks ranked by the excess return prediction to form up the portfolio for the next trading day. In order to reduce the investment risk, traders usually distribute the investing equally over these top-$k$ stocks. Thus, $k$ is a hyper-parameter to balance between the robustness and profits of the portfolio. In our experiments, we set up $4$ sets of comparative experiments with $k$ to be $30$, $50$, $100$ and $200$ respectively. To further approximate the real-world trading, we consider a transaction cost of $0.4\%$ for both buy and sell transactions. Actual accumulated returns are used to evaluate the trading policies under these different strategies.

The backtesting results are given in Figure \ref{fig6}. Under different values of $k$, our ADD method achieves higher accumulated income than the baseline models (GRU baseline models). Over time, the income gap between the baseline models and our ADD is increasing gradually, indicating that our method is stable and appropriate along time. Our method has obvious advantages, no matter when the prediction effect is better or when the performance is not that good. Through backtesting, we verify that ADD leads to a promotion in the actual investment that can have real practical value.

\section{Related Work}

\subsection{Disentanglement Representation}

Disentanglement representation extraction aims to figure out explanatory factors of the input data to generate more meaningful features. Early attempts including bilinear model \cite{tenenbaum2000separating}, autoencoder \cite{georghiades2001few}, manifold learning \cite{elgammal2004separating}, Restricted Boltzmann Machine \cite{reed2014learning} and so on. Recently, Variational Autoencoder (VAE) and Generative Adversarial Networks (GANs) are found to be more powerful. Besides, researchers are also interested in explorations about the ability of disentanglement representation to solve challenge tasks \cite{bengio2013representation,van2019disentangled,locatello2019fairness}. 

The disentanglement method is first introduced to stock prediction in \cite{hadad2018two}, which proposes a two-step disentanglement method by training the two encoders separately. Unlike its asynchronous disentanglement, our disentanglement process is fully end-to-end and trained synchronously. Besides, our novel framework and training phases differ from their method a lot. 

\subsection{Knowledge Distillation}
Knowledge distillation usually utilizes a teacher-student strategy, which targets at transferring the knowledge from a strong teacher model to a student model \cite{bucilua2006model,ba2014deep,hinton2015distilling}. Previous purpose of knowledge distillation lies on model compression, which intends to reduce the large model space of a teacher model to a small student model while trying best to keep the performance. Techniques include soft label matching \cite{hinton2015distilling}, adding random perturbations into soft label \cite{sau2016deep}, hidden layer approximating \cite{romero2014fitnets} and attention weight mapping \cite{zagoruyko2016paying}.

The performance of student model is always worse than the teacher model, self-distillation is proposed to achieve comparable or even better performance than teacher model, which distills the knowledge between teacher and student models in an identical architecture \cite{yim2017gift,furlanello2018born,zhang2019your,kim2020self}. We adopt self-distillation but differ from them in two key points. First, we propose a dynamic self-distillation process by introducing some weights to leverage the knowledge from teacher model and the information in raw data. Second, our distillation method is performed with disentanglement network, which not only improves the model performance, but also assists in eliminating interference factors.

\section{Conclusion}
Stock trend forecasting plays a more attractive role nowadays with the help of deep learning approaches. In this work, we propose an Augmented Disentanglement Distillation (ADD) approach to remove the inferential features from the noised raw stock data, so as to extract clean information and improve the market/excess return prediction performances. Our ADD contains a disentanglement structure, a dynamic self-distillation method, as well as a novel data augmentation strategy. Experiments on the Chinese stock market data demonstrate the effectiveness of our approach for stock trend forecasting. Also, the backtesting experiments also verify that ADD has real impact in the actual investment to help traders increase income. For future works, we are interested in simplifying the training strategy as well as designing a more effective disentanglement structure to improve stock trend forecasting. 

\bibliographystyle{neurips_2020}
\bibliography{neurips_2020}

\begin{thebibliography}{36}
\providecommand{\natexlab}[1]{#1}
\providecommand{\url}[1]{\texttt{#1}}
\expandafter\ifx\csname urlstyle\endcsname\relax
  \providecommand{\doi}[1]{doi: #1}\else
  \providecommand{\doi}{doi: \begingroup \urlstyle{rm}\Url}\fi

\bibitem[Ba \& Caruana(2014)Ba and Caruana]{ba2014deep}
Jimmy Ba and Rich Caruana.
\newblock Do deep nets really need to be deep?
\newblock In \emph{Advances in neural information processing systems}, pp.\
  2654--2662, 2014.

\bibitem[Bengio et~al.(2013)Bengio, Courville, and
  Vincent]{bengio2013representation}
Yoshua Bengio, Aaron Courville, and Pascal Vincent.
\newblock Representation learning: A review and new perspectives.
\newblock \emph{IEEE transactions on pattern analysis and machine
  intelligence}, 35\penalty0 (8):\penalty0 1798--1828, 2013.

\bibitem[Bisoi \& Dash(2014)Bisoi and Dash]{bisoi2014hybrid}
Ranjeeta Bisoi and Pradipta~K Dash.
\newblock A hybrid evolutionary dynamic neural network for stock market trend
  analysis and prediction using unscented kalman filter.
\newblock \emph{Applied Soft Computing}, 19:\penalty0 41--56, 2014.

\bibitem[Bucilua et~al.(2006)Bucilua, Caruana, and
  Niculescu-Mizil]{bucilua2006model}
C~Bucilua, R~Caruana, and A~Niculescu-Mizil.
\newblock Model compression, in proceedings of the 12 th acm sigkdd
  international conference on knowledge discovery and data mining, 2006.

\bibitem[Chen et~al.(2019)Chen, Zhao, Bian, Xing, and Liu]{chen2019investment}
Chi Chen, Li~Zhao, Jiang Bian, Chunxiao Xing, and Tie-Yan Liu.
\newblock Investment behaviors can tell what inside: Exploring stock intrinsic
  properties for stock trend prediction.
\newblock In \emph{Proceedings of the 25th ACM SIGKDD International Conference
  on Knowledge Discovery \& Data Mining}, pp.\  2376--2384, 2019.

\bibitem[Chen et~al.(2015)Chen, Zhou, and Dai]{chen2015lstm}
Kai Chen, Yi~Zhou, and Fangyan Dai.
\newblock A lstm-based method for stock returns prediction: A case study of
  china stock market.
\newblock In \emph{2015 IEEE international conference on big data (big data)},
  pp.\  2823--2824. IEEE, 2015.

\bibitem[Chung et~al.(2014)Chung, Gulcehre, Cho, and
  Bengio]{chung2014empirical}
Junyoung Chung, Caglar Gulcehre, KyungHyun Cho, and Yoshua Bengio.
\newblock Empirical evaluation of gated recurrent neural networks on sequence
  modeling.
\newblock \emph{arXiv preprint arXiv:1412.3555}, 2014.

\bibitem[Elgammal \& Lee(2004)Elgammal and Lee]{elgammal2004separating}
Ahmed Elgammal and Chan-Su Lee.
\newblock Separating style and content on a nonlinear manifold.
\newblock In \emph{Proceedings of the 2004 IEEE Computer Society Conference on
  Computer Vision and Pattern Recognition, 2004. CVPR 2004.}, volume~1, pp.\
  I--I. IEEE, 2004.

\bibitem[Feng et~al.(2018)Feng, Chen, He, Ding, Sun, and
  Chua]{feng2018enhancing}
Fuli Feng, Huimin Chen, Xiangnan He, Ji~Ding, Maosong Sun, and Tat-Seng Chua.
\newblock Enhancing stock movement prediction with adversarial training.
\newblock \emph{arXiv preprint arXiv:1810.09936}, 2018.

\bibitem[Furlanello et~al.(2018)Furlanello, Lipton, Tschannen, Itti, and
  Anandkumar]{furlanello2018born}
Tommaso Furlanello, Zachary~C Lipton, Michael Tschannen, Laurent Itti, and
  Anima Anandkumar.
\newblock Born again neural networks.
\newblock \emph{arXiv preprint arXiv:1805.04770}, 2018.

\bibitem[Georghiades et~al.(2001)Georghiades, Belhumeur, and
  Kriegman]{georghiades2001few}
Athinodoros~S. Georghiades, Peter~N. Belhumeur, and David~J. Kriegman.
\newblock From few to many: Illumination cone models for face recognition under
  variable lighting and pose.
\newblock \emph{IEEE transactions on pattern analysis and machine
  intelligence}, 23\penalty0 (6):\penalty0 643--660, 2001.

\bibitem[Hadad et~al.(2018)Hadad, Wolf, and Shahar]{hadad2018two}
Naama Hadad, Lior Wolf, and Moni Shahar.
\newblock A two-step disentanglement method.
\newblock In \emph{Proceedings of the IEEE Conference on Computer Vision and
  Pattern Recognition}, pp.\  772--780, 2018.

\bibitem[Hinton et~al.(2015)Hinton, Vinyals, and Dean]{hinton2015distilling}
Geoffrey Hinton, Oriol Vinyals, and Jeff Dean.
\newblock Distilling the knowledge in a neural network.
\newblock \emph{arXiv preprint arXiv:1503.02531}, 2015.

\bibitem[Hochreiter \& Schmidhuber(1997)Hochreiter and
  Schmidhuber]{hochreiter1997long}
Sepp Hochreiter and J{\"u}rgen Schmidhuber.
\newblock Long short-term memory.
\newblock \emph{Neural computation}, 9\penalty0 (8):\penalty0 1735--1780, 1997.

\bibitem[Hu et~al.(2018)Hu, Liu, Bian, Liu, and Liu]{hu2018listening}
Ziniu Hu, Weiqing Liu, Jiang Bian, Xuanzhe Liu, and Tie-Yan Liu.
\newblock Listening to chaotic whispers: A deep learning framework for
  news-oriented stock trend prediction.
\newblock In \emph{Proceedings of the eleventh ACM international conference on
  web search and data mining}, pp.\  261--269, 2018.

\bibitem[Ioffe \& Szegedy(2015)Ioffe and Szegedy]{ioffe2015batch}
Sergey Ioffe and Christian Szegedy.
\newblock Batch normalization: Accelerating deep network training by reducing
  internal covariate shift.
\newblock \emph{arXiv preprint arXiv:1502.03167}, 2015.

\bibitem[Kim et~al.(2020)Kim, Ji, Yoon, and Hwang]{kim2020self}
Kyungyul Kim, ByeongMoon Ji, Doyoung Yoon, and Sangheum Hwang.
\newblock Self-knowledge distillation: A simple way for better generalization.
\newblock \emph{arXiv preprint arXiv:2006.12000}, 2020.

\bibitem[Kingma \& Ba(2014)Kingma and Ba]{kingma2014adam}
Diederik~P Kingma and Jimmy Ba.
\newblock Adam: A method for stochastic optimization.
\newblock \emph{arXiv preprint arXiv:1412.6980}, 2014.

\bibitem[Li et~al.(2019)Li, Song, and Tao]{li2019multi}
Chang Li, Dongjin Song, and Dacheng Tao.
\newblock Multi-task recurrent neural networks and higher-order markov random
  fields for stock price movement prediction: Multi-task rnn and higer-order
  mrfs for stock price classification.
\newblock In \emph{Proceedings of the 25th ACM SIGKDD International Conference
  on Knowledge Discovery \& Data Mining}, pp.\  1141--1151, 2019.

\bibitem[Li et~al.(2016)Li, Leng, Yang, and Yu]{li2016stock}
Lili Li, Shan Leng, Jun Yang, and Mei Yu.
\newblock Stock market autoregressive dynamics: A multinational comparative
  study with quantile regression.
\newblock \emph{Mathematical Problems in Engineering}, 2016, 2016.

\bibitem[Locatello et~al.(2019)Locatello, Abbati, Rainforth, Bauer,
  Sch{\"o}lkopf, and Bachem]{locatello2019fairness}
Francesco Locatello, Gabriele Abbati, Thomas Rainforth, Stefan Bauer, Bernhard
  Sch{\"o}lkopf, and Olivier Bachem.
\newblock On the fairness of disentangled representations.
\newblock In \emph{Advances in Neural Information Processing Systems}, pp.\
  14611--14624, 2019.

\bibitem[McWilliams \& Siegel(1997)McWilliams and Siegel]{mcwilliams1997event}
Abagail McWilliams and Donald Siegel.
\newblock Event studies in management research: Theoretical and empirical
  issues.
\newblock \emph{Academy of management journal}, 40\penalty0 (3):\penalty0
  626--657, 1997.

\bibitem[Metghalchi et~al.(2012)Metghalchi, Marcucci, and
  Chang]{metghalchi2012moving}
Massoud Metghalchi, Juri Marcucci, and Yung-Ho Chang.
\newblock Are moving average trading rules profitable? evidence from the
  european stock markets.
\newblock \emph{Applied Economics}, 44\penalty0 (12):\penalty0 1539--1559,
  2012.

\bibitem[Minh et~al.(2018)Minh, Sadeghi-Niaraki, Huy, Min, and
  Moon]{minh2018deep}
Dang~Lien Minh, Abolghasem Sadeghi-Niaraki, Huynh~Duc Huy, Kyungbok Min, and
  Hyeonjoon Moon.
\newblock Deep learning approach for short-term stock trends prediction based
  on two-stream gated recurrent unit network.
\newblock \emph{Ieee Access}, 6:\penalty0 55392--55404, 2018.

\bibitem[Nelson et~al.(2017)Nelson, Pereira, and de~Oliveira]{nelson2017stock}
David~MQ Nelson, Adriano~CM Pereira, and Renato~A de~Oliveira.
\newblock Stock market's price movement prediction with lstm neural networks.
\newblock In \emph{2017 International joint conference on neural networks
  (IJCNN)}, pp.\  1419--1426. IEEE, 2017.

\bibitem[Reed et~al.(2014)Reed, Sohn, Zhang, and Lee]{reed2014learning}
Scott Reed, Kihyuk Sohn, Yuting Zhang, and Honglak Lee.
\newblock Learning to disentangle factors of variation with manifold
  interaction.
\newblock In \emph{International Conference on Machine Learning}, pp.\
  1431--1439, 2014.

\bibitem[Romero et~al.(2014)Romero, Ballas, Kahou, Chassang, Gatta, and
  Bengio]{romero2014fitnets}
Adriana Romero, Nicolas Ballas, Samira~Ebrahimi Kahou, Antoine Chassang, Carlo
  Gatta, and Yoshua Bengio.
\newblock Fitnets: Hints for thin deep nets.
\newblock \emph{arXiv preprint arXiv:1412.6550}, 2014.

\bibitem[Sau \& Balasubramanian(2016)Sau and Balasubramanian]{sau2016deep}
Bharat~Bhusan Sau and Vineeth~N Balasubramanian.
\newblock Deep model compression: Distilling knowledge from noisy teachers.
\newblock \emph{arXiv preprint arXiv:1610.09650}, 2016.

\bibitem[Srivastava et~al.(2014)Srivastava, Hinton, Krizhevsky, Sutskever, and
  Salakhutdinov]{srivastava2014dropout}
Nitish Srivastava, Geoffrey Hinton, Alex Krizhevsky, Ilya Sutskever, and Ruslan
  Salakhutdinov.
\newblock Dropout: a simple way to prevent neural networks from overfitting.
\newblock \emph{The journal of machine learning research}, 15\penalty0
  (1):\penalty0 1929--1958, 2014.

\bibitem[Tenenbaum \& Freeman(2000)Tenenbaum and
  Freeman]{tenenbaum2000separating}
Joshua~B Tenenbaum and William~T Freeman.
\newblock Separating style and content with bilinear models.
\newblock \emph{Neural computation}, 12\penalty0 (6):\penalty0 1247--1283,
  2000.

\bibitem[van Steenkiste et~al.(2019)van Steenkiste, Locatello, Schmidhuber, and
  Bachem]{van2019disentangled}
Sjoerd van Steenkiste, Francesco Locatello, J{\"u}rgen Schmidhuber, and Olivier
  Bachem.
\newblock Are disentangled representations helpful for abstract visual
  reasoning?
\newblock In \emph{Advances in Neural Information Processing Systems}, pp.\
  14245--14258, 2019.

\bibitem[Wei(2016)]{wei2016hybrid}
Liang-Ying Wei.
\newblock A hybrid anfis model based on empirical mode decomposition for stock
  time series forecasting.
\newblock \emph{Applied Soft Computing}, 42:\penalty0 368--376, 2016.

\bibitem[Xu \& Cohen(2018)Xu and Cohen]{xu2018stock}
Yumo Xu and Shay~B Cohen.
\newblock Stock movement prediction from tweets and historical prices.
\newblock In \emph{Proceedings of the 56th Annual Meeting of the Association
  for Computational Linguistics (Volume 1: Long Papers)}, pp.\  1970--1979,
  2018.

\bibitem[Yim et~al.(2017)Yim, Joo, Bae, and Kim]{yim2017gift}
Junho Yim, Donggyu Joo, Jihoon Bae, and Junmo Kim.
\newblock A gift from knowledge distillation: Fast optimization, network
  minimization and transfer learning.
\newblock In \emph{Proceedings of the IEEE Conference on Computer Vision and
  Pattern Recognition}, pp.\  4133--4141, 2017.

\bibitem[Zagoruyko \& Komodakis(2016)Zagoruyko and
  Komodakis]{zagoruyko2016paying}
Sergey Zagoruyko and Nikos Komodakis.
\newblock Paying more attention to attention: Improving the performance of
  convolutional neural networks via attention transfer.
\newblock \emph{arXiv preprint arXiv:1612.03928}, 2016.

\bibitem[Zhang et~al.(2019)Zhang, Song, Gao, Chen, Bao, and Ma]{zhang2019your}
Linfeng Zhang, Jiebo Song, Anni Gao, Jingwei Chen, Chenglong Bao, and Kaisheng
  Ma.
\newblock Be your own teacher: Improve the performance of convolutional neural
  networks via self distillation.
\newblock In \emph{Proceedings of the IEEE International Conference on Computer
  Vision}, pp.\  3713--3722, 2019.

\end{thebibliography}

\end{document}